\documentclass[12pt,acmtog,noacm]{acmart}
\pdfoutput=1 
\AtBeginDocument{%
  \providecommand\BibTeX{{%
    \normalfont B\kern-0.5em{\scshape i\kern-0.25em b}\kern-0.8em\TeX}}}

\title{Using Navigational Information to Learn Visual Representations}\thanks{This extended abstract was presented at the {\it Computational and Systems Neuroscience (COSYNE) Conference}, Lisbon and Cascais, Portugal, March 2022.}\thanks{Some computation of this work used the Extreme Science and Engineering Discovery Environment (XSEDE), which is supported by National Science Foundation grant number ACI-1548562. Discussions with Sitao Zhang were helpful. We are also grateful for the support of the ThreeDWorlds development team and the assistance of Justin Lee.}
\author{Lizhen Zhu, Brad Wyble, James Z. Wang}
\affiliation{%
  \institution{The Pennsylvania State University}
  \city{University Park}
  \state{Pennsylvania}
  \postcode{16802}
  \country{USA}
}
\email{{ljz5180,bpw10,jwang}@psu.edu}

\renewcommand\footnotetextcopyrightpermission[1]{} 
\begin{document}
\maketitle

\textbf{Abstract.} Children learn to build a visual representation of the world from unsupervised exploration and we hypothesize that a key part of this learning ability is the use of self-generated navigational information as a similarity label to drive a learning objective for self-supervised learning. The goal of this work is to exploit navigational information in a visual environment to provide performance in training that exceeds the state-of-the-art self-supervised training.  Here, we show that using spatial and temporal information in the pretraining stage of contrastive learning can improve the performance of downstream classification relative to conventional contrastive learning approaches that use instance discrimination to discriminate between two alterations of the same image or two different images. We designed a pipeline to generate egocentric-vision images from a photorealistic ray-tracing environment (ThreeDWorld) and record relevant navigational information for each image. Modifying the Momentum Contrast (MoCo) model, we introduced spatial and temporal information to evaluate the similarity of two views in the pretraining stage instead of instance discrimination. This work reveals the effectiveness and efficiency of contextual information for improving representation learning. The work informs our understanding of the means by which children might learn to see the world without external supervision.

\vspace{-0.1in}
\section{Introduction and motivation}
The ability to learn a robust representation from the visual statistics of reflected light that permits action and perception is a fundamentally critical property of biological minds that we are still far from understanding. However, the visual experience of a child in the first year of life is extremely limited compared to datasets used in modern computer vision approaches and they are still able to learn quite effectively. Many children spend their first year of life largely within a single house populated by a fairly limited and largely unchanging set of furnishings and objects. Within this first year, they will make approximately 40 million visual fixations which roughly constitutes the number of views they experience. In contrast, modern image sets for self-supervised training have over a billion images with many object types. Yet, human infants, despite the limited visual input and their lack of ability to understand verbal labels, lay the foundation of an extremely robust visual representation that is to be matched by algorithms. We test the hypothesis that spatiotemporal contiguity provides a form of ground truth inherent in hippocampal representations of time~\cite{howard2002distributed} and space~\cite{o1976place} that can be used for efficient self-supervised learning in a limited visual environment.   

\vspace{-0.12in}
\section{Egocentric-vision image generation}
We generate our dataset on the ThreeDWorld platform~\cite{gan2020threedworld}, which provides flexibility to change the environment, the objects in the environment, the lighting conditions, and many other parameters. In the 3D environment, an avatar is controlled to move along a certain trajectory which can be designed by humans in advance or generated algorithmically. The stored trajectory can be used to generate the image samples from one or more lighting conditions.

\vspace{-0.12in}
\section{Contrastive learning using spatiotemporal context }

Contrastive learning allows self-supervised learning by training an encoder with instance discrimination. However, this algorithm ignores the similarity between images. In contrast, human perception has continuity, which could make use of the changes of perspective, lighting conditions, and other information to learn invariant representations across realistic physical changes in an image, rather than the more limited artificial augmentations used in modern training algorithms. We posit that even very young children possess the ability to track spatial and temporal information by virtue of hippocampal representations, which provides the basis for similarity tagging. No other sources of similarity information are assumed.


Our model is based on MoCo ~\cite{chen2020mocov2}. An image $i$ is randomly augmented into two views, which are encoded separately by two encoders with the same structure to obtain feature $q_i$ and feature $k_i$.  $k_i$, as well as related information, are stored in a fixed-sized dictionary. Multiplying two features, the similarity of the two views can be obtained. The loss function is a variant of infoNCE:
$$L=-\sum \log\frac{\exp(q_i\cdot k_{j(i)}/\tau)}{\sum_{a\in A(i)}\exp(q_i\cdot k_a/\tau)}\;.$$
Here, $\tau$ is a temperature parameter, and $j(i)$ denotes the index of the views which are positive pairs with $i$. $A(i)$ is the set that includes all the negative pairs with $i$. Instead of labeling the views generated from the same image as positive and those from different images as negative, we propose that the similarity is related to spatial and temporal information of images. 

For a given image, we choose a positive image that falls within a certain threshold of steps (for the time objective, Time MoCo) or location and rotational angle (for the spatial objective, space MoCo) to pair with it. Images outside thresholds are considered negative pairs. We choose only one image rather than all to stay as close to the Standard MoCo implementation as possible, which used one positive pair for each instance.   

\begin{figure}[t]
    \centering
    \includegraphics[width=3.5in]{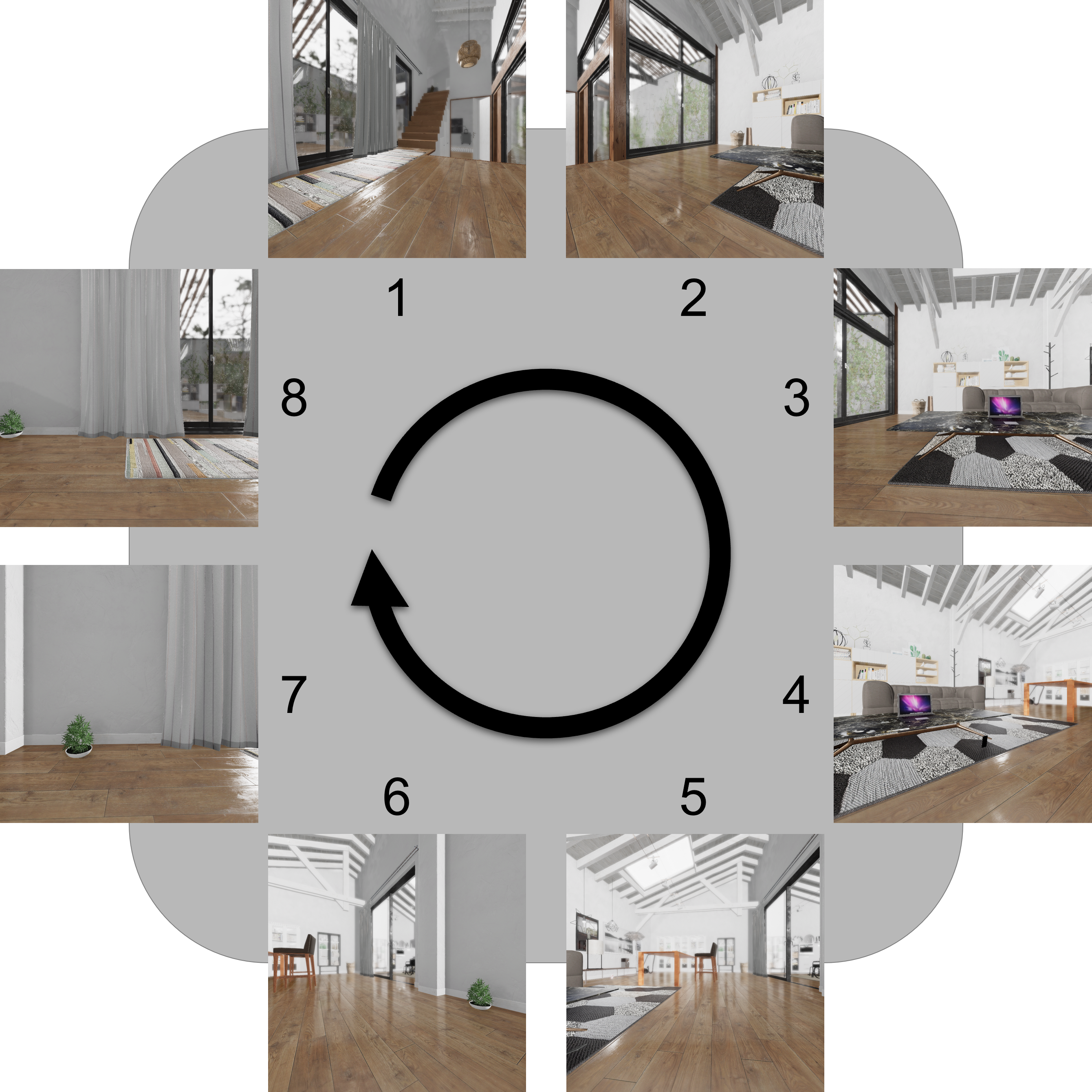}
    \caption{Example images from a short trajectory using our pipeline as the agent turns in a circle. Images that are close in space or time are considered similar.}
\label{fig:data}
\end{figure}

\vspace{-0.12in}
\section{Experiment and results}
In the experiment, we used a continuous trajectory that explored the indoor environment of a house in ThreeDWorld, archviz, that was furnished with household objects from the ThreeDWorld model library. The dataset that we collect includes $14,767$ high-quality images from a single trajectory through the house that was recorded by a human using keyboard controls that enabled movement in four cardinal directions, as well as horizontal rotation and vertical jumps (Fig.~\ref{fig:data}). A data file records the timestamp, rotation, position for an image collected at each step through the house. Images were downsized to $224\times 224$ pixels. 

\begin{table}[H]
    \centering
    \caption{Result of Standard MoCo, Time MoCo and Space MoCo on ImageNet dataset. Accuracies are the average of runs in percentage. $N$ represents the number of times the model was run.}
    \label{table}
    \begin{tabular}{cccc}
    \toprule[2pt]
    Model&N&ImNet top-1 acc.& Std Dev\\\hline
    Standard MoCo& 2 &14.26  & 0.57 \\
    Time MoCo &2 &15.00 & 0.47\\
    Space MoCo&2 &\textbf{16.57} & 0.19\\\hline
    Standard MoCo&5&4.13 & 0.50\\
    Space MoCo &5&\textbf{7.17} & 0.81\\
    \bottomrule[2pt]
    \end{tabular}
\end{table}
We compare Time MoCo and Space MoCo with Standard MoCo in Table~\ref{table} with an extended regime consisting of $800$ epochs of pretraining on our dataset followed by $200$ of downstream training on Imagenet.  The thresholds for similarity in time, position, and rotation are $10$ seconds, $0.2$ meter, and $3$ degrees, respectively. Overall accuracy is low because the backbone is trained only on images from the house, but this allows a fair comparison of our similarity training and the instance discrimination of MoCo.  
The data shows that Space MoCo reliably outperforms Standard MoCo. Time MoCo gets a worse result than Space MoCo, for it ignores the similarity of images obtained from similar positions and rotations but at a different time. For example, in Fig.~\ref{fig:data},
images 1 and 8 might be considered as a negative pair by Time MoCo, but as a positive pair by Space MoCo.

To further demonstrate this reliability we re-ran Standard and Space MoCo five times each using a shorter regime with 200 epochs pre-training and 50 epochs downstream training and got similar results. 

\section{Future work}
Several directions can be taken to extend this work. Contrastive training can be extended to larger generated datasets to learn better representations. Morever, because the labels needed in our model are relative spatiotemporal information, the training can be modified to use sensor data to compute spatial and temporal proximity rather than using ground truth.
The potential applications of this new learning concept will be well beyond the ImageNet classification tasks we have tested.



\bibliographystyle{ACM-Reference-Format}
\bibliography{ref}

\end{document}